\documentclass{article} 
\usepackage{nips15submit_e,times}
\usepackage[cmex10]{amsmath}
\usepackage{amssymb}
\usepackage{amsmath}
\usepackage{dsfont}
\usepackage{amsthm}
\usepackage{bm}
\usepackage{mathtools}
\usepackage{times}
\usepackage{enumerate}
\usepackage[noadjust]{cite}
\usepackage{algorithm2e}
\usepackage{graphicx}
\usepackage{svg}
\usepackage{calc}
\usepackage{moreverb}
\usepackage{paralist}
\usepackage{caption}
\usepackage{booktabs}
\usepackage{import}
\usepackage{tikz}
\usetikzlibrary{shapes,shadows}
  \tikzstyle{abstractbox} = [draw=black, fill=white, rectangle, 
  inner sep=10pt, style=rounded corners, drop shadow={fill=black,
  opacity=1}]
  \tikzstyle{abstracttitle} =[fill=white]

\newtheorem{theorem}{Theorem}[section]
\newtheorem{lemma}[theorem]{Lemma}
\newtheorem{axiom}{Axiom}[section]

\newtheorem{proposition}[theorem]{Proposition}
\newtheorem{corollary}[theorem]{Corollary}

\theoremstyle{definition}
\newtheorem{definition}{Definition}[section]

\theoremstyle{definition}
\newtheorem{example}{Example}[section]

\makeatletter
\newenvironment{proof*}[1][\proofname]{\par
  \pushQED{\qed}%
  \normalfont \partopsep=\z@skip \topsep=\z@skip
  \trivlist
  \item[\hskip\labelsep
        \itshape
    #1\@addpunct{.}]\ignorespaces
}{%
  \popQED\endtrivlist\@endpefalse
}
\makeatother

\newcommand\numberthis{\addtocounter{equation}{1}\tag{\theequation}}

\newcommand{\Conv}{\mathop{\scalebox{1.5}{\raisebox{-0.2ex}{$\ast$}}}}%

\numberwithin{equation}{section}

\newif\ifcompilefull
\compilefullfalse 

\title{(Yet) Another Theoretical Model of Thinking}

\author{
Patrick Virie\\
p.virie@gmail.com
}

\nipsfinalcopy 

\begin{document}

\maketitle

\begin{abstract}

	This paper presents a theoretical, idealized model of the thinking process with the following characteristics: 
	\begin{inparaenum}[1)]
		\item the model can produce complex thought sequences and can be generalized to new inputs,
		\item it can receive and maintain input information indefinitely for the generation of thoughts and later use, and
		\item it supports learning while executing.
	\end{inparaenum}
	The crux of the model lies within the concept of \emph{internal consistency}, 
	or the generated thoughts should always be consistent with the inputs from which they are created.
	Its merit, apart from the capability to generate new creative thoughts from an internal mechanism,
	depends on the potential to help training to generalize better. 
	This is consequently enabled by separating input information into several parts 
	to be handled by different processing components with a focus mechanism to fetch information for each.
	This modularized view with the focus binds the model with the computationally capable Turing machines.
	And as a final remark, this paper constructively shows that the computational complexity of the model 
	is at least, if not surpass, that of a universal Turing machine.
	
\end{abstract}

\section{Introduction}

This paper presents a theoretical, idealized model of thinking. 
Thinking, as a mental process, is one of the most sophisticate products of intelligence.
It allows us to perform procedural simulations in order to predict the future outcomes of given present states.
Having the model of thinking would enable the explanation of our mind and would also facilitate the building of replicas of it.

Due to its deep inherent association with our minds, thinking is one of the fundamental concepts in philosophy \cite{boole1854investigation}. 
Perhaps most of the modern attempts to model the process of thinking are arguably inspired by Alan Turing's work \cite{turing1950computing}, 
which addresses many philosophical and technical questions about the machines that can think.
From that point onward, the term "thinking" has been associated with various meanings even within the context of computation. 
Our interpretation of the word thinking would only limited to the continuous process of generating data from selected inputs.

As the reader goes through the content, a question might come up in mind, "this does not seem to be how the brain works." 
The purpose of this paper however, is not to postulate the actual process of thinking that happens in one's brain; 
we are only interested a theoretical fraction that captures the ideal essence of thinking such as 
how new ideas are composed from experience and generalization of thinking sequences.
From this motive, the model that we develop in this paper only has to maintain the following characteristics:
\begin{inparaenum}[a)]
\item it must permit complex transformation within sequence and generalize to unseen inputs,
\item it can receive and maintain information indefinitely and selectively use them to generate future sequences, and
\item it should support learning while executing.
\end{inparaenum}
From which necessities are these characteristics derived? And how do we address them? 

If we manage to represent the information of any moment of thought using a structure within a mathematically defined space,
thinking process can be defined as a sequence of transformations between the structures within that space.
To be able to capture real world sequences, the transformations should be sufficiently expressive.
This problem can recently be addressed by the re-popularized concept of deep learning \cite{hinton2006fast, bengio2009learning, schmidhuber2015deep}. 
Deep learning allows stacking of simple transformations, which are individually easy to analyze, to express a complex non-linear transformation.

Even though thoughts are very fluid and alternating from moment to moment, yet everything happens in mind.
Every information that we generate or receive at some points in time shall be used at some other times to generate a new data.
Therefore to address the second characteristic, the model must be able to memorize information.
The success of the recurrent models for time-related tasks shows that we might use them as a prototype of our model \cite{hochreiter1997long, graves2009novel}. 
This will be discussed in Section \ref{section_model}.
Furthermore to show that the complexity of the model could rival that of our mind, 
we could try to relate the model with the behavior of a universal Turing machine \cite{hennie1966two}.
Section \ref{section_turing} will contribute to this aim.

Finally, we aim to develop a model that allows learning process to happen homogeneously along the execution path and simultaneously at the execution time. 
This requirement is not crucial in the process of thinking,
but it is useful for any actual system that implements our model to have some kind of online-learning capability. 
We will show that learning while executing is possible within our model in Section \ref{section_model}.

Before going to the model, we will first discuss the constraint that constitutes the essence of our model, called \emph{internal consistency}. 
We will relate this constraint with the notion of generalization within the scope of thinking.
What can we guarantee given the inputs can change? How would we define generalization in the context of thought?
And how can we implement the constraint in a real machine?
This is where we start.

 
\section{Internal consistency}
\label{section_internal_consistency}
	
	\begin{figure}[t]
		\centering
		\def\svgwidth{0.3\textwidth}
		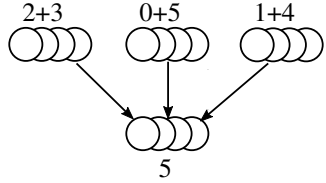
		\caption{Alternative representations of ``5''}
	\end{figure}	

	For a generative system that tries to model the world, it usually consists of two sides of data representation: 
	one that represents the observed, visible states of the world, and the other that represents the hidden states \cite{koller2009probabilistic}. 
	The hidden states are \emph{alternative representations} of the visible states, and as the name implies, 
	each represents another way of how to represent a data. 
	For example, we say $2+3$ is an alternative representation of $5$, 
	so do $1 + 4$ and any other summation of two numbers that equal $5$.

	Why do we have to be bothered with alternative representations?
	Alternative representation systems usually have some neat inter-basis characteristics.
	For example, the bases that contain information in the hidden states can be made less correlated, less redundant, or sometimes be completely independent.
	Independent bases are desirable for generative models, since we can represent the distributions of the world efficiently as the products of distributions of individual bases \cite{hyvarinen2000independent}: $P(X_0, X_1, \ldots) = \prod_i P(X_i)$ where $X_i$s are random variables.
	It is up to applications to define the best sets of bases for the hidden states. 
	This is in fact conformed to what deep learning tries to achieve. 

	To further illustrate why the alternative representations are necessary as the fundamental of our thinking model.  
	What if whenever we try to imagine the Mona Liza but what comes out from our mind instead turns out to be the Scream. 
	Is it not frustrating? Though these two historical paintings are of artistic merits, 
	recalling a wrong one while we originally intend for the other can hardly be recognized as a trait of intelligence without proper reasons.	 
	In control theory, engineers attempt to design systems with closed-loop sensory feedbacks to rectify undesirable outputs.
	The systems can immediately identify when their previous outputs deviate from the expectation in order to compute controls with proper compensation schemes. 
	Goodfellow \emph{et al.} suggested a neat mathematically-proven training technique to regularize this behavior in neural networks by introducing adversarial networks \cite{goodfellow2014generative}. The adversarial networks will attempt to identify whether data are the results from generative networks or the real distributions. The generative ones must be trained to counter this discrimination as best as they could.
	Despite the success of these techniques, would it not be better if we can build systems that can inherently prevent all of these action-intention inconsistencies?
	We can say that the systems are perfectly adaptable without the need for compensatory countermeasures. 
	In fact, this is an empirical trait that any intelligent systems should have, something that we humans have, at least to some degrees.

	Let us suppose for now that the inputs of a generative system that tries to generate visible states of the world are never-been-seen-before hidden states.
	How can we guarantee that the system would produce the correct visible states as the outputs?
	The generalization of alternative representations would be achieved by making sure that the generated visible representations are always conformed to the hidden states that cause them. For example, suppose that we have a system that tries to produce the addition result of any two numbers, say $A$ and $B$ as the hidden state, we say the system is generalized when its generative function is $A+B$.
	We call the phenomenon where the hidden states of a system are always alternative representations of its visible states \emph{internal consistency}. 
	This is the best we can do for generalization. 

	What does this heuristic mean computationally? 


	\subsection{The non-sharing property and preservation of variance}

	Let $v$ represents a visible state from the set $V$, and $h$ represents a hidden state from the set $H$.
	A system is said to be internally consistent when the mapping from a visible state to any hidden state 
	and the reconstruction from one of the reachable hidden states always results in the visible state itself:
	\begin{align}
		\sum_h P(v|h)P(h|v) = 1 \,\,\, \forall v \in V
	\end{align}
	Or the mapping preserves variance in $V$.	
	Please note that this paper uses probabilistic short forms; namely, $P(h|v)$ is a short form of $P(\hat{h}=h|\hat{v}=v)$ where $\hat{h}, \hat{v}$ are random variables.

	To better understand the connection between internal consistency and preservation of variance,
	we need to consider the fact that an alternative representation of any data may not be unique. 
	A visible state can have many alternative hidden states of representation, 
	but those states must only correspond to none other than the visible state itself.
	\begin{definition}
		Let a set of alternative hidden states of a visible state is a set whereas every element can be generatively mapped into the visible state.
		The \emph{non-sharing property} is satisfied if and only if the forward mapping from the visible state only results in an element from the set.
	\end{definition}
	The non-sharing property implies that, when we transform a hidden state into its corresponding visible state, 
	there is no other visible state that better matches the hidden state available.
	Let us think about the addition example again. Suppose that we want to produce the visible state of $2+3$, which is $5$, 
	how do we guarantee that the produced $5$ is correct? It is by converting back $5$ into a hidden state, which may result in $1+4$. Then 
	we verify that $1+4$ and $2+3$ belong to the same set. 


	\begin{figure}[t]
		\centering
		\def\svgwidth{1.0\textwidth}
		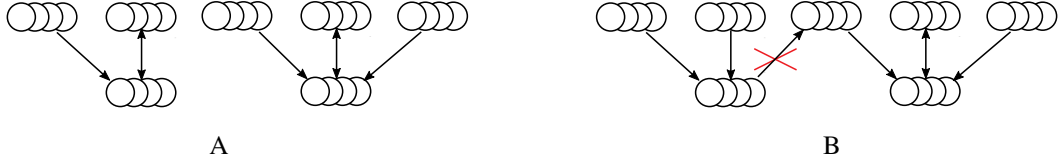
		\caption{The non-sharing property. The group on the left has the non-sharing property. The group on the right does not. The cross-mark indicates which mapping is not valid.}
	\end{figure}	

	\begin{lemma}
		Preservation of variance implies the non-sharing property.
		\label{lemma_variance_to_non_sharing}
		\begin{proof*}
			$\sum_h P(h|v) P(v|h)$ is a convex combination of $P(v|h)$ $\forall v,h$.
			If there exists $P(v|h) < 1$ then $\sum_h P(h|v) P(v|h) < 1$.
			Therefore, for any $h$ with non-zero $P(h|v)$, $P(v|h) = 1$.
		\end{proof*}	
	\end{lemma}

	\begin{lemma}
		The non-sharing property is equivalent to variance preservation.
		\label{lemma_variance_preserve}
		\begin{proof*}
			For any $h$ with non-zero $P(h|v)$, $\sum_h P(v|h)P(h|v) = 1$ when $P(v|h) = 1$, $\forall v$.
			Also, from Lemma \ref{lemma_variance_to_non_sharing}. This completes the proof.
		\end{proof*}
	\end{lemma}		

	From Lemma \ref{lemma_variance_preserve}, for a system to be internally consistent, it must at least preserve the variance in the visible states.
	Internal consistency may not be perfectly achieved in practice. We can however approach it by gradually training the system to maximize reconstruction chance:
	\begin{align}
		\max \sum_h P(v|h)P(h|v)  \,\,\, \forall v \label{relax}
	\end{align}
	Learning hidden representation while maximizing the reconstruction chance aligns with the goal of autoencoder training \cite{bengio2009learning, vincent2010stacked}. Preservation of variance is therefore another justification for learning representation with autoencoders.		
	
	One nice property of the non-sharing property is that it can be stacked to create a deep expressive architecture that permits multi-layer hidden data transformation. This way, it is possible to find a good complex representation for any data domain by having each layer gradually removes correlation in the data and promotes little-by-little independency for the data in the adjacent layer.
	Training to build a deep internally consistent system is as simple as training to preserve the variance between layers.

	\begin{lemma}
		Stacking of variance preservation satisfies the non-sharing property. \label{stack_adaptive_recon}%
		\begin{proof*}
			Let the prime notations of one dimension vectors represent their variants.
			Given a variance preservation layer, $\sum_h P(v|h)P(h|v) = 1$, stacking another layer on top of it preserves the non-sharing property:
			$\sum_h P(v|h) \left( \sum_{h'} P(h|h') P(h'|h) \right) P(h|v) = \sum_h P(v|h) 1 P(h|v) = 1$.
			We can apply this action multiple times to build a multi-layer architecture. This completes the induction. 
		\end{proof*}
	\end{lemma}	

	Preservation of variance suggests a way to generate innovative yet relevant visible states from new hidden states,
	and therefore allows us to guarantee internal consistency, i.e., alternative representation for unseen inputs.
	To show this in a system, we require the knowledge or the detail of system implementation.
	We will see in the next section a way to implement an internally consistent system.

	\subsection{Implementation in a linear system}

	We discuss a linear neural network as an internally consistent system.
	Each layer of the network can be mathematically expressed as a matrix multiplication: 
	$h = Wv$ where $W$ represents a forward linear mapping weight matrix, $v$ is a visible state vector, and $h$ is a hidden state vector. 
	To achieve variance preservation, the generative mapping from a hidden state to a visible state $W'$ must fulfills this equation: $v = W'Wv$. 
	For the analytical purpose, it is even simpler to consider filling the entire linear span of the visible state set, i.e., to make $W'W = I$.
	In this regard, the generative matrix $W'$ must be the left inverse of the weight matrix $W$, thus preserving variance of the visible states.
	The following subsection shows that we can guarantee the non-sharing property for unseen visible states when the inverse exists.		
	
	\subsubsection{A linear neural layer that satisfies internal consistency}

		Let us define the concept of equilibrium in a linear neural layer.
		\begin{definition}
			An equilibrium is a setting where a visible state and and only one of its hidden states correspond to each other. 
		\end{definition}
		In other words, for any $h \in H_v$ or the set of alternative hidden representations of $v$, every $P(v|h) = 1$, 
		and only an element $\hat{h} \in H_v$ receives all the probabilistic mass, $P(\hat{h}|v) = 1$. 
		For short notation, $\hat{h} \iff v$.
		This is partly due to the determinism of linear systems that allows no more than one $h$ to correspond to $Wv$ 
		and only one $v$ to reciprocally correspond to $W'h$.
		\begin{lemma}
			A linear system that allows hidden-visible transformation to reach the equilibrium in one step from any hidden state has the non-sharing property.
			\label{single_step_lemma}
			\begin{proof*}
				To prove this statement, we show that, when the non-sharing property does not hold, there always exists a path greater than one step.
				Let $v' \to h \to v$ be a path. Such a part violates the non-sharing property when $v' \neq v$. Suppose there exists an $h'$ such that
				$h' \to v$. $h'$ can never be $h$, because $h \to v$ and $h \to v'$ cannot be true at the same time; this contradicts the determinism of linear systems. From here we can conclude that there is at least more than one step from $h'$ to reach the closest equilibrium, $h' \to v' \to h \iff v$.
			\end{proof*}
		\end{lemma}

		\begin{proposition}
			A linear system where its forward mapping has the left inverse satisfies the non-sharing property for unseen hidden states.
			\begin{proof*}
				Consider a path $h' \to v' \to h'' \to v'' \to \ldots$ in its linear form:
				$h' \to W'h' \to WW'h' \to W'WW'h' \to \ldots$
				Since $W'W = I$, the path can be truncated:
				$h' \to W'h' \to WW'h' \to IW'h' = W'h' \iff WW'h'$.
				From any $h'$, the system reaches the equilibrium $W'h' \iff WW'h'$ in one step.
				Thus, it satisfies the non-sharing property according to Lemma \ref{single_step_lemma}.
			\end{proof*}
		\end{proposition}

		\begin{corollary}
			Stacking of linear internally consistent systems satisfies the non-sharing property for unseen hidden states.
			\begin{proof*}
				This is true by Proposition \ref{stack_adaptive_recon}.
				It can also be seen that when each layer satisfies the non-sharing property, 
				any forward pass in the stack is always an equilibrium path.
				Since the stack only takes one single generative pass from a hidden state to generate its visible state,
				it satisfies the non-sharing property according to Lemma \ref{single_step_lemma}.
			\end{proof*}
		\end{corollary}

	\subsubsection{Missing variance}

		The discussion of alternative representation has led us to a kind of problems 
		where in some applications we might want to construct visible states purely from given hidden states, i.e., constructing $v$ from $h$.
		Many real life problems belong to such category, to state a few, 
		giving an artistic style and an abstract shape, how to create a detailed painting containing the shape with the style \cite{szegedy2014going, gatys2015neural} or reconstructing visual images from brain reading signals \cite{nishimoto2011reconstructing}.
		The difficulty of these problems, apart from finding the generative mapping, lies mostly in the fact that 
		it is nearly impossible to perfectly provide all the variance for the generative construction.
		The construction requires that all the variance in the hidden state have to be filled; 
		otherwise the result visible state may suffer the lack of sufficient details. 
		This might coincide why even for us it is sometimes hard to imagine the precise details of some concepts. 
		Before discussing how could we fill the missing variance, we need a way to represent it first.

		We introduce a \emph{hidden residue} as a part of a hidden state that is not given as an input.	
		Let $W$ represents an assumed-given visible-to-hidden mapping weight, 
		and $U$ represents a weight for visible-to-residue mapping. 
		To satisfy internal consistency, we must have the left inverse $W'|U'$ of $W|U$
		such that 
		\begin{align}
				(W'|U')(W|U)v = (W' W + U' U)v = v 	\label{objective_internal_consistency}
		\end{align}
		$A|B$ is a matrix as a concatenation result between the matrix $A$ and $B$ of the same number of columns. 

		We can constructing $v$ from $h$ via this relationship:
		\begin{align}
				v = W'h + U'r \label{feed_down_residue}
		\end{align}
		where $r$ is the have-to-be-inferred hidden residue state.
		The process of inference must be automatically done by the system.
		In the next section, we explore the strategies to train a system with such capability 
		following the internal consistency constraint.

	\begin{figure}[t]
		\centering
		\def\svgwidth{0.3\textwidth}
		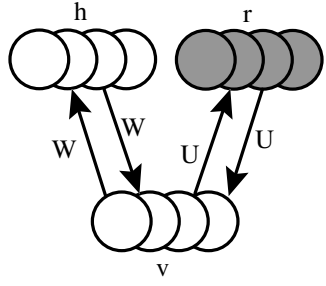
		\caption{An internally consistent model with a hidden residue state representation.}
	\end{figure}	

	\subsection{Training internal consistency in a linear system}

		The goal of training is to search for a set of bases that provides good alternative representations and preserves variance in the data.
		This is unsupervised learning. Given a set of visible states $v \in V$, we wish to find the forward mapping weight $W$, its generative weight $W'$, also perhaps along with the residue mapping $U$, and its inverse $U'$, subject to Equation \ref{objective_internal_consistency}.

		Despite that the exact solution for finding the weights can be found via any decomposition process,
		it is more favorable in applications with large amount of data to use an iterative based algorithm.
		Reconstruction ICA is a good candidate for training \cite{le2011ica}.
		Consider one form of its objective (without the residue weight),
		$\text{min}_W \sum_{v \in \text{training data}} \left( ||W^\intercal W v - v ||^2_2 + \lambda || Wv ||_1 \right)$,
		Le \emph{et al.} shows that
		\begin{lemma}[Le \emph{et al.}, 2011]
			The reconstruction term in RICA's objective is equivalent to $|| (W^\intercal W - I)E\Lambda^{\frac{1}{2}} ||^2_\mathcal{F}$ 
			or the orthonormality cost with weights in the space rotated by eigenvectors and scaled by eigenvalues.
		\end{lemma}
		$\Lambda$ is a diagonal eigenvalue matrix, and $E$ is a matrix whose columns are eigenvectors of the covariance matrix of the training data.
		$\lambda$ in RICA's objective is a sparsity coefficient that controls how much learning effort contributes to finding independent bases.
		When the eigenvalues of the training data are real positive, we can see that the solution to RICA involves ones where $W^\intercal W$ approaches the identity, which in turn makes the system that implements RICA satisfies internal consistency. 

		Although RICA is originally derived with the generative weight as the transpose of the forward weight and without the residue weights, 
		we can extend it to support a non-transpose system by updating the gradients for both $W$ and $W'$ separately, and expand the weight into the form $W|U$, which includes the residue weight.

		\subsubsection{Linear transpose bases}

		In deep learning, it has been a common approach to use the transpose of the forward weight as the generative one.
		The transpose acts as a good learning regulator originally presented in Oja rule's \cite{oja1982simplified}. 

		For a linear neural layer with transposes, the weights that suffice the condition for internal consistency follow 
		\begin{align*}
			v &= \left(W^\intercal W + U^\intercal U\right) v
			\intertext{which also implies that}
			U^\intercal U v &= (I - W^\intercal W) v \\
			W^\intercal W v &= (I - U^\intercal U) v
			\intertext{When $v$ is any vector from the entire linear span of the visible state set, we can see that}
			U^\intercal U &= (I - W^\intercal W) \\
			W^\intercal W &= (I - U^\intercal U)
		\end{align*}
		Unless we allow complex values in the neural network, 
		the last two equations suggest that only the weights that make $I - W^\intercal W$ and $I - U^\intercal U$ 
		have real positive eigenvalues can be decomposed into $U^\intercal U$ and $W^\intercal W$ respectively.
		If the weights are valid and the square roots of their eigenvalues exist, 
		knowing one weight allows us to immediately extract the other weight via any eigen decomposition process of the form
		\begin{align*}
			U^\intercal U &= I - W^\intercal W = E \Lambda E^\intercal \\
			U &= \Lambda^{\frac{1}{2}} E^\intercal
		\end{align*}
		$\Lambda$ is a diagonal eigenvalue matrix, and $E$ is a matrix whose columns are eigenvectors of $I - W^\intercal W$,
		and vice versa for $W$.

		Since each eigenvalue represents the variance along each principle axis in the visible state set, 
		this means each of the weights, $W$ and $U$, cannot extract more information than that presenting in the visible state set.

		\subsubsection{Addressing missing variance}
		\label{section_memory_implementation}
		
		When some parts of the hidden states are not given, we can fill them with the priors from training.
		This suggests a system with some form of internal memory that can remember the information provided in the hidden states during the training.
		The memory allows the system to later infer the residue states conditioned on the available part of the given hidden states.
		
		We choose to fulfill the role of the memory with a belief network stacked on top of the hidden units 
		due to its simplicity \cite{hinton2006fast} among other generative techniques.	
		A belief network is a generative model that can be trained to generate training data's distribution $P(v)$ where $ v \in $ the training set.
		It is a stack of restricted Boltzmann machines trained with the contrastive divergence algorithm, 
		a variant of gradient ascent with the following update rule: $\nabla W \propto \sum_v P(v) \left( \sum_h P(h|v) hv^\intercal - \sum_{h',v'} P(h',v') h'v'^\intercal \right)$. 
		At convergence, the values of the hidden units $h$ conditioned on the visible units' $v$ should match 
		the stationary distribution due to bipartite nature of the machines.
		Inference in a belief network creates a Markov chain with its stationary distribution conforms to $P(v)$.

		In our case, we use the top layer belief network to model the distribution of the hidden and the residue state, i.e., $v$ of the belief network is simply $h|r$ of our system. Given $h$, running the inference in the form of repetitive sampling should provide us the missing variance conditioned on it.
		Then the hidden state and the residue state are fed down to construct the visible state following Equation \ref{feed_down_residue}.

		We will later see that this variance filling mechanism can be used as a memory extension for our thinking model. 
		This augmentation allows us to increase to the model's capacity to quickly access memory and also facilitates the learning while executing procedure.


	\begin{figure}[h!]
		\centering
		\def\svgwidth{0.3\textwidth}
		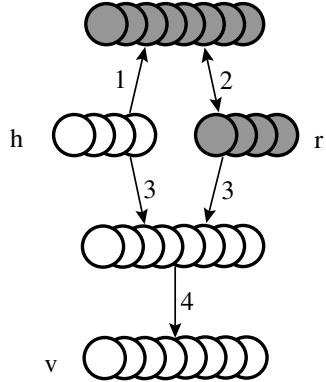
		\caption{A stack of internally consistent layers with its inference steps to recover the missing variance. 
		The blacken figures indicate a belief network with $r$ as the hidden residue state representation of this model.}
		\label{figure_stacks}
	\end{figure}

	\subsection{Extension to non-linear}

		In some applications, having hidden states that capture some non-linear traits in visible states has never been without practical advantages.
		We can use the rectified linear activation function as a means to achieve this \cite{nair2010rectified}.
		The non-linear behavior of rectified linear units can convey non-linear information from visible states to hidden states while satisfying internal consistency given the presence of \emph{mirror bases}.

		\begin{lemma}
			A rectified linear layer with mirror bases satisfies internal consistency in the same manner to those with linear bases.
			\begin{proof*}
				Let $Wu$ be represented by $\rho(Wu) + \rho(-Wu)$ when $\rho$ is a rectified linear projection of every row of $Wu$. 
				Let $k, u$ be column vectors of the same length:
					\begin{align*}
						\rho(k^\intercal u) = \begin{dcases*}
									        k^\intercal u  & when $k^\intercal u > 0$\\
									        0 & otherwise
									 \end{dcases*}
					\end{align*}
				Let $W'$ represents the left inverse of $W$.
				It can be seen that for any $v$, $W'\rho(Wv) + -W'\rho(-Wv) = W' Wv = v$. 
			\end{proof*}
			\label{lemma_rect_mirror}
		\end{lemma}
		
		A pair of mirror bases contains mutually exclusive linear polars tied together to form a linear basis.
		Although within a layer, each mirror pair behaves like a linear basis and follows internal consistency,
		we can have non-linear transfer at the adjacent layer by assigning a different weight value to each of the rectified linear base in the pair.


		\begin{figure}[h!]
			\centering
			\def\svgwidth{0.7\textwidth}
			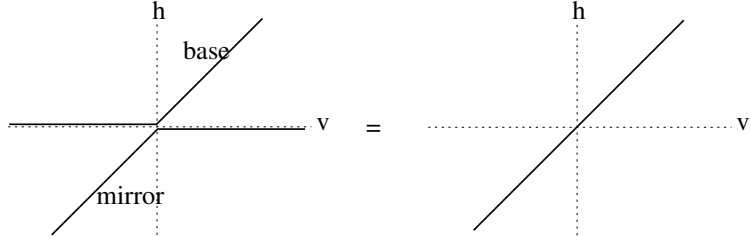
			\caption{A rectified linear mirror pair is equivalent to a linear function.}
			\label{figure_mirror_base}
		\end{figure}

		\subsubsection{A more general way to choose non-linear bases}
		\label{general_nonlinear}
		
		Here we suggest a method to modify a linear mapping into a non-linear one that, of course, satisfies internal consistency.
		
		Let us start from the basic linear internal consistency in a linear layer.
		\begin{align*}
			\intertext{For any $v$,}
			v 	&= 	W'Wv \\
				&=	W'IWv \\
				&=	W' \begin{bmatrix} I & I & \ldots \end{bmatrix} 
						\begin{bmatrix} S_0 \\ S_1 \\ \ldots \end{bmatrix} Wv
			\intertext{The last one requires that}
			 &\sum_i S_i 	= 	I 	\numberthis \label{identity_spectrum}
		\end{align*}
		Note that each basis set $S_i$ is not only a constant matrix but can also be a collection of functions 
		where their resolved values depend on to what values they are being multiplied. 		
		\begin{example}
			Let $\sigma$ represents a step function at $0$:
			\begin{align*}
				\sigma(x).x &= \begin{cases}
					1.x = x & \mbox{if } x \geq 0 \\
					0.x = 0 & \mbox{otherwise}
				\end{cases}
				\intertext{We can see that}
				\sigma(x).x &= \rho(x)
			\end{align*}	
			The product of the step function with its input behaves like a rectified linear function,  
			and the mirror pair of $\sigma$ is in fact $1 - \sigma$.
			With this, we have an option to choose basis sets as follow:
			\begin{align*}
				S_0 &= \begin{bmatrix} \sigma & 0 & \ldots \\ 0 & \sigma & \ldots \\ \ldots & \ldots & \ldots \end{bmatrix}\\
				S_1 &= \begin{bmatrix} 1-\sigma & 0 & \ldots \\ 0 & 1-\sigma & \ldots \\ \ldots & \ldots & \ldots \end{bmatrix}
			\end{align*}
			which satisfies Equation \ref{identity_spectrum}.

		\end{example}

		For a layer, we can then expressively have
		\begin{align*}
			h 	&=	\begin{bmatrix} S_0 \\ S_1 \\ \ldots \end{bmatrix} Wv \numberthis \label{pre_activate_linear} \\
			v 	&=	W' \begin{bmatrix} I & I & \ldots \end{bmatrix} h\text{.}
		\end{align*}
		Note that the formed generative and forward mapping are not symmetry. 
		The generative mapping is just a summation before being multiplied with the left inverse 
		while the forward activation is non-linear following the choice of basis sets.

		To allow non-linear transfer across layers, the adjacent layer can assign a different weight value to each basis. 
		We can use this fact with a finite number of basis functions to represent almost any bounded complex transformation according to the universal approximation theorem \cite{cybenko1989approximation}.

		As a theoretical remark, we can also see that the infinite number of local basis functions can approximate any transformation in fact:
		\begin{align*}
			 f(x) &=  \int^{\infty}_{0^+} \frac{f(\hat{x})}{\hat{x}}.\pi_{\hat{x}}(x).x \, \mathrm{d}\hat{x} + \int^{0^-}_{-\infty} \frac{f(\hat{x})}{\hat{x}}.\pi_{\hat{x}}(x).x \, \mathrm{d}\hat{x}
		\end{align*}
		where
		\begin{align*}
			\pi_{\hat{x}}(x)  &= \begin{cases}
						1 & \mbox{if } x = \hat{x} \\
						0 & \mbox{otherwise}
					\end{cases}
		\end{align*}
		If we treat $\frac{f(\hat{x})}{\hat{x}}$ as the adjacent layer weight for the basis that corresponds to $\hat{x}$ value, 
		we can regard $\pi_{\hat{x}}$ as the basis itself. The following equation holds:
		\begin{align*}
			 y \int^{\infty}_{-\infty} \pi_{\hat{x}}(x).x \, \mathrm{d}\hat{x}  &= y.1.x \quad \forall y,x
		\end{align*}
		In the multiplicative case however, there is an only exception for the critical point at 
		$x = 0$, and there is no information to be transferred.
		This can be leveraged with a tweak to both the forward and the generative mapping to detect 
		the critical point and assign a unique basis value to allow correct inversion.

		\begin{figure}[t]
			\centering
			\def\svgwidth{1.0\textwidth}
			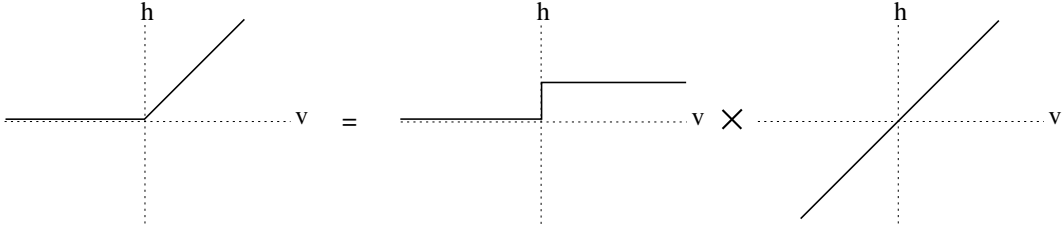
			\caption{A rectified linear can be factorized into a step function and a linear function.}
			\label{figure_non_linear}
		\end{figure}

		The application of this proposal is that we can perform the following steps when training internal consistency in a neural layer:
		\begin{enumerate}
			\item Train its linear weight matrices following $v = W'Wv$
			\item Choose non-linear basis sets $ \begin{bmatrix} S_0 \\ S_1 \\ \ldots \end{bmatrix} $ such that $\sum_i S_i = I$
		\end{enumerate}
		This technique can be applied many times in a stack to form alternating layers of linear and non-linear transfer functions.

		A variation of the technique can be used in convolutional networks \cite{atlas1988artificial} as well:
		\begin{align*}
			v 	&=	W' \Conv W \Conv v \\
			v 	&=	W' \Conv \delta \Conv W \Conv v \\
			v 	&=	W' \Conv \left\{ \begin{matrix} S_0 \\ S_1 \\ \ldots \end{matrix} \right\} \Conv W \Conv v  \numberthis \label{conv_nonlinear}
		\end{align*}
		$\delta$ is the Dirac delta function, the identity of the convolution operation.
		For a short notation we define the following quantity $\left\{ \begin{matrix} S_0 \\ S_1 \\ \ldots \end{matrix} \right\}$ 
		an alternative representation of the identity $\mathcal{I}$ that belongs to any tensor operation $\star$ with the distributive property such that
		\begin{align*}
			A \star \left\{ \begin{matrix} S_0 \\ S_1 \\ \ldots \end{matrix} \right\} \star B &= A \star \mathcal{I} \star B \\
			\sum_i S_i &= \mathcal{I}
		\end{align*}
		$A$ and $B$ belong to the domain of the $\star$ operation.

		From Equation \ref{conv_nonlinear} we can have
		\begin{align*}		
			h_i	&=	 S_i \Conv W \Conv v 				\numberthis \label{pre_activate} \\						
			v	&=	W' \Conv \sum_i h_i 							
		\end{align*}

		In practice however, it is not convenient to convolve a tensor with a basis set that depends on the value of its operand as in Equation \ref{pre_activate}. 
		We can leverage this with the convolution theorem and the inner product:
		\begin{align*}
			v 	&=	\mathcal{F}' (\mathcal{F} (W') \cdot \mathcal{F} ( \mathcal{F}' ( \mathcal{F}(W) \cdot \mathcal{F}(v) ) ) ) \\
			v 	&=	\mathcal{F}' (\mathcal{F} (W') \cdot  \mathcal{F}(W) \cdot \mathcal{F}(v) ) )  \\
			v 	&=	\mathcal{F}' (\mathcal{F} (W') \cdot \mathds{1} \cdot \mathcal{F}(W) \cdot \mathcal{F}(v) ) )  \\
			v 	&=	\mathcal{F}' (\mathcal{F} (W') \cdot \left\{ \begin{matrix} S_0 \\ S_1 \\ \ldots \end{matrix} \right\} \cdot \mathcal{F}(W) \cdot \mathcal{F}(v) ) )
		\end{align*}
		$\mathcal{F}$ and $\mathcal{F}'$ are the Fourier transform and its inverse respectively.

		We can now multiply each element individually as in the linear example:
		\begin{align*}
			h_i 	&=  \mathcal{F}'(S_i \cdot \mathcal{F}(W) \cdot \mathcal{F}(v) ) 						\numberthis
		\end{align*}

		Again, we can first train the convolution kernels $W$ and $W'$ before choosing the non-linear transfer functions.

	\subsection{Temporal internal consistency}
	\label{temporal_section}

		Recurrent neural networks belong to a class of expressive architectures 
		that we can use to build generative temporal systems for modeling temporal sequences \cite{taylor2006modeling}.
		If we treat the hidden states as the \emph{theme} of a temporal sequence, 
		we can have the visible states that can change through time
		by treating past states as the condition for the present time step.
		For each past configuration, the current visible state will satisfy the non-sharing property.
		But does the non-sharing property apply for unseen conditions as well?

	\begin{figure}[ht!]
		\centering
		\def\svgwidth{0.5\textwidth}
		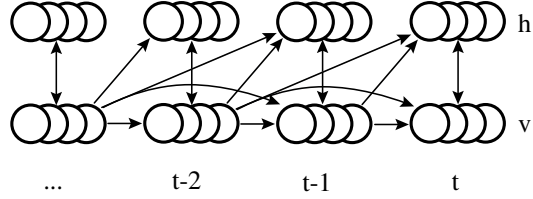
		\caption{ One layer of a temporal model conditioned on two past steps. }
		\label{figure_temporal_model}
	\end{figure}

		Let $v_t$ represents the visible state at a time $t$. 
		$\vec{v}_{\setminus t}$ denotes a collection of visible states over many time steps into some past excluding $v_t$.
		Consider the expectation of temporal variance preservation conditioned on $\vec{v}_{\setminus t}$:
		$\sum_{\vec{v}_{\setminus t}} P(\vec{v}_{\setminus t}|v_t) \sum_h P(v_t | h, \vec{v}_{\setminus t}) P(h| v_t, \vec{v}_{\setminus t}) = 1$.	
		We can rearrange the term to treat the past data as a part of the hidden state; 
		and together with the original hidden state, we can use them to produce the visible state of the present time step:
		\begin{align*}
			\sum_{\vec{v}_{\setminus t}} \left[ \sum_h P(v_t | \vec{v}_{\setminus t}, h) P(h| \vec{v}_{\setminus t}, v_t) \right] P(\vec{v}_{\setminus t}|v_t) = 
			\sum_{\vec{v}_{\setminus t}} \sum_h P(v_t | \vec{v}_{\setminus t}, h) P(h, \vec{v}_{\setminus t} | v_t) = 1
		\end{align*}
		\begin{lemma}
			\label{temporal_adaptive_recon}
			Following the proof of Lemma \ref{lemma_variance_to_non_sharing}, 
			satisfying the conditional non-sharing property is equivalent to satisfying the non-sharing property 
			when the condition is accounted as a part of the hidden state. 
		\end{lemma}%
		Therefore in any conditional alternative representation system, we can consider the conditions as a part of the hidden states.
		As time progresses, each visible state will always be alternatively represented by the combination of a hidden state and some past visible states, satisfying the non-sharing property for any condition and thus the internal consistency constraint.

		To build a temporal internally consistent system, 
		the training algorithm has to preserve the variance found in the present visible state conditioned on the given past visible states.
		Due to the relation from Lemma \ref{temporal_adaptive_recon}, we propose another proposition:
		\begin{proposition}
		If an algorithm that is used to train the conditional non-sharing property shares the same routines with one 
		that is used to train the non-sharing property when the condition is accounted as a part of the hidden state,
		that algorithm will generalize the non-sharing property to unseen conditions.
		\end{proposition}
		It remains depending on the system implementation to provide sufficient criteria that can guarantee the generality of the algorithm to unseen conditions.
		Analogous to the non-conditional case, it is straightforward to see that conditional RICA can be regarded as such an algorithm.

		We can also extend the expressions to describe multilayer systems:
		\begin{align*}
			&\sum_{\vec{v}^0_{\setminus t}} \sum_{v^1_t} P(v^0_t | \vec{v}^0_{\setminus t}, v^1_t) \left( \sum_{\vec{v}^1_{\setminus t}} \sum_{v^2_t}  P(v^1_t | \vec{v}^1_{\setminus t}, v^2_t) \left( \ldots \right) P(v^2_t, \vec{v}^1_{\setminus t} | v^1_t) \right)  P(v^1_t, \vec{v}^0_{\setminus t} | v^0_t) =\\
			&\sum_{\vec{v}^0_{\setminus t}} \left[ \sum_{v^1_t} P(v^0_t | \vec{v}^0_{\setminus t}, v^1_t) \left( \sum_{\vec{v}^1_{\setminus t}} \sum_{v^2_t}  P(v^1_t | \vec{v}^1_{\setminus t}, v^2_t) \left( \ldots \right) P(v^2_t, \vec{v}^1_{\setminus t} | v^1_t) \right)  P(v^1_t | \vec{v}^0_{\setminus t} , v^0_t) \right]  P(\vec{v}^0_{\setminus t} | v^0_t)
		\end{align*}
		The superscripts denote layers to which the associated states belong.

		The temporal model allows visible states to change despite having been corresponded to a fixed hidden state.
		The hidden states can also change following the temporal progression of the visible states.
		We could stop here and propose these two alternative transformation as the model of thinking 
		unless we wish to integrate the concept of attention into our thinking paradigm.
	
\section{Yet another model of thinking}
\label{section_model}
	
	
	Our model of thinking is a temporal apparatus that continually generates data
	to form a sequence of thoughts using the information from somewheres and some times in the sequence itself.
	In a manner similar to other alternative representation models, we represent the generated thoughts with visible states,
	and we use hidden states to relate information in the thought sequence. 
	The hidden states are divided and grouped forming a finite number of processing modules which we call components.
	Each component acts as an information cache whose content is fetched depending on a spatio-temporal cue, or simply \emph{focus}.
	Components are used to provide the means to choose information from various sources
	and combine them in a creative yet deducible way to form the sequence of thinking.

	While recurrent neural networks are useful,
	their recurrent connections have a disadvantage compared to focuses.
	To generate a thought using recurrent connections that requires some information up to some long past in the sequence,
	the recurrent connections have to cover throughout a lot of past steps in the sequence.
	But this would come into a shortcoming when we consider our minds' capability to switch between different thoughts.
	At one time, a song is just an ear-worm in our mind,
	but at another time, we would have no problem to switch to another song. 
	If it is the extensive temporal recurrent connections we have in our mind, switching thoughts would not be so simple. 
	The cue that signals the switching would have to compete with many others.
	This idea suggests that the model should permit only short temporal connections, 
	and should rather rely on another mechanism to fetch information, such as the focus.
	With the focus, the model can access past information at any time while allowing the ability to abruptly change thoughts.

\begin{figure}[ht!]
		\centering
		\def\svgwidth{0.75\textwidth}
		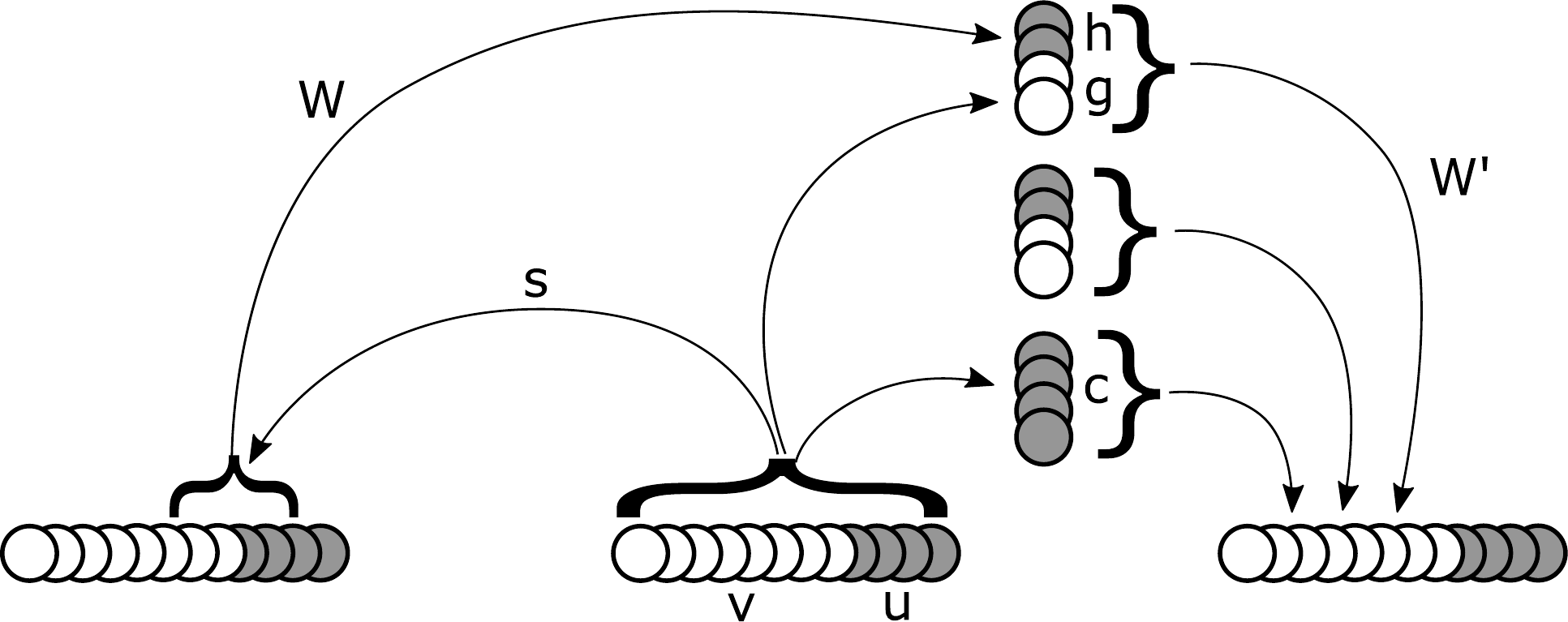
		\caption{An illustration of our model. 
		Here we have three visible states depicted horizontally at the bottom.
		From the middle state which is our current step, the model infers a selective and a generative focus $s$ and $g$. 
		$s$ points to a location in the past and the information from that location is fetched 
		and transformed via a forward mapping weight $W$ to one of the components
		depicted vertically. The result hidden states along with the generative weight $W'$ are used to produce a new visible state.
		$c$ is a context component.
		$u$ is an augmented external input as a part of a visible state representation.}
		\label{the_model}
	\end{figure}

	The impression of the model is best illustrated visually in Figure \ref{the_model}.

	\subsection{The model and its behavior}
			
	In our model,
	the degrees of freedom that change between consecutive thought steps derive only from focuses.
	In particular, a focus controls from where and when information should be fetched into a component.
	A focus is similar to a pointer in the context of programming, namely, an address that points to a memory content in a program.
	Though the memory content can be dynamically altered, the part of the program that executes and controls the pointer remained static.
	It is this concept that we bring into our model with focuses, and this allows us to generalize the model to some degree.
	The focus also allows the model to choose which portion of information should and shall be processed at a time 
	yielding a biologically inspired capability to recognize structures in noisy environments if any.
	
	Let us define the focus of a component as a tuple of a selective focus $s$ and a generative focus $g$, $f = (s,g)$. 
	The selective focus defines from where in the previous visible states the information content in the component is fetched, 
	and the generative focus defines upon which portion of the new visible state the information content will be placed.

	At the beginning of each time step $t$, the model starts by activating the focus of each component using only the recent thoughts.	
	\begin{equation} \label{focus}
		v_{t}, v_{t-1}, f_{t-1}, \ldots, v_{t-\tau}, f_{t-\tau}  \to f = (s, g)  \to h
	\end{equation}
	The activation of the focus is unidirectional from a finite number $\tau$ of latest visible states in the sequence,
	with optionally the same number of previous focuses, to the current focus of each component. 
	The lack of a backward mapping means the non-sharing property between the visible states and the focus is not defined. 
	Yet when the activation is deterministic, the non-sharing property is implicitly hold because
	we can always find a backward mapping that can sustain it.

	Once the focuses of every component are generated, 
	the model then fetches the content $h$ for each component, corresponding to each individual's selective focus $s$,
	and uses it together with the contents from other components to generate a new visible state, 
	a new thought for the thought sequence, following the internal consistency constraint. 
	
	\begin{equation} \label{combine}
    	\forall v \sum_{\pmb{h}} P(v | \pmb{h}, \pmb{g}) P(\pmb{h} | v, \pmb{g^{-1}}) = 1 
    \end{equation}
    We call this the combination rule, i.e., the generated thought must preserve all the information provided in the components.
    $\pmb{h}$ represents a collection of all components' contents.
    $\pmb{g}$ is a collection of generative focuses from all components that dictates how should $\pmb{h}$ be combined.
    In order to complete the equation, we require the inverses of the generative focuses $\pmb{g^{-1}}$. 
    They act as the selective focuses that choose information from the newly generated thought back into the components.  

    Since the generative focuses of the components may sometimes not overlap one another.
	To make the combination rule always valid, the content mapping parameters of each individual component must satisfy preservation of variance. 
	For each component $i$,
	\begin{equation} \label{detail}
    	\forall v_{i} \sum_{h_i} P(v_{i}| h_i ) P( h_i| v_{i} ) = 1
    \end{equation}	
	This equation serves as a constraint that explains the behavior of the mapping parameters for each individual component;
	it is not mandated to hold for every portion of the generated thought; 
	there can be times when this equation contradicts Equation \ref{combine}, e.g., 
	when the generative focuses place the mutually contradictory contents of two or more components on the same portion of the thought.
	Nevertheless, it must hold for the mapping parameters of each component.

   	All of these equations constitute the thinking process of our model.
   		
   	\subsection{Advantages}

   		For the reader, it is best to pause here and discuss the advantages this model can accomplish.

		First, the multi component model allows each component to store information from a different source 
		and be ready to combine with others' to form a new thought.
		As we mention in the previous section that when the model has been trained to follow internal consistency, 
		we could have a generalization guarantee for the alternative representation to any new combination.
		Creativity is therefore governed by the focuses of the components that select information, 
		and from which allows new thoughts to form.
		
		Sir Isaac Newton had postulated in \emph{his seminal work} \cite{newton1999principia} 
		that every surrounding change in the environment comes from either motion or transformation. 
		Some objects may alter their intrinsic properties, but the rest only move.
		Using our model, it is possible to have a representation of any environment state that separates motion from the background,
		and we let the focuses handle the mechanic of the moving part. This way we aim for a better generalization when training the model.
		
	 	The last reason is in fact a means to reduce hypothesis variance with a limited amount of training data.
		To see roughly why separating focuses and contents can help reduce hypothesis variance,
		we can count the amount of training examples required for two neural network implementations, i.e., with and without focuses.
		Let $M$ be the number of our model's components,
		$T$ be the total number of past states we kept for the network model without focuses,
		$N$ be the number of content bits per each state,
		$K$ be the number of values per bit,
		Finally we let each individual bit of focus has two values, i.e., focus or not focus.
		In the worst case scenario where the networks can only memorize the input-output pairs and do not generalize them,
		the lower-bound of the required number of examples to memorize input-output mappings is the size of domain times the size of co-domain.
		For the network model without focus, the required number of examples is the size of total past states we kept times the size of the new state, $T(K^N) \times K^N$.	We will let the reader work out for the required number of examples for our model, which is $M \left( K^{N/M} \times K^{N/M} \right) + (K^N \times 2^N)$.
		We can see that for case $M = 2$, $K \geq 2$, and $N \geq 2 $;
		the amount of examples for training the model without focuses is greater than the number required for our model.
		Usually for a fair comparison we let $T = M$, i.e., 
		the number of states we kept for the network model without focuses is equal to the number of components of our model.
		\begin{align*}
			T(K^N) \times K^N &= TK^{2N} \\ 
							  &= 2K^{2N} \\
							  &> M \left( K^{N/M} \times K^{N/M} \right) + (K^N \times 2^N) \\
							  &= M K^{2N/M} + (2K)^N \\
							  &= (2^N + 1)K^N
		\end{align*}
		This crude estimation only provides an intuition of why separating focuses can help reducing hypothesis variance of a neural network however, 
		as we do not take account of the ability to generalize of the compared models nor the dependencies in the training data. 


	\subsection{Existence}

		Here we show that in general, we can always build a multicomponent-multilayer internally consistent system 
		that allows non-linear representation of the visible state sequences following our model's behavior.
		\begin{proposition}
			There exist non-linear implementations of the model that satisfy Equation \ref{combine} and Equation \ref{detail}.
			\label{lemma_combination}
		\end{proposition}
		Giving some examples will take care of the proof.
		\begin{example}
			In the first example, we show that a linear, single-layer, internally consistent implementation of the model exists. 
			Then by Lemma \ref{stack_adaptive_recon} and Lemma \ref{lemma_rect_mirror}, 
			we can put any desired number of non-linear layers at the bottom of it to create a stack of non-linear internal consistency layers.
	
			From Equation \ref{combine}, we interpret it into a linear form:
				\begin{align}			
					v = \bm{G} \bm{W}' \bm{W} \bm{G}^{-1} v	\quad \forall v \label{linear_combine}
				\end{align}
			which can be expanded as the example below:
				\begin{align*}
					v = 
					\begin{bmatrix}
						G_{0} &		
						G_{1} &		
						G_{2}
					\end{bmatrix}				
					\begin{bmatrix}
						W'_{0} & 0 & 0 \\		
						0 & W'_{1} & 0 \\		
						0 & 0 & W'_{2}
					\end{bmatrix}
					\begin{bmatrix}
						W_{0} & 0 & 0 \\		
						0 & W_{1} & 0 \\		
						0 & 0 & W_{2}
					\end{bmatrix}	
					\begin{bmatrix}
						G^{-1}_{0} \\		
						G^{-1}_{1} \\		
						G^{-1}_{2}
					\end{bmatrix}							
					v	
				\end{align*}
			$G_i$ and $G^{-1}_i$ are the generative focus matrix of a component $i$ and its inverse respectively.
			$W_i$ and $W'_i$ are the forward content mapping matrix of a component $i$ and the corresponding generative mapping matrix.
			To satisfy Equation \ref{detail}, we require that 
				\begin{align}
					x = W'_i W_i x \quad \text{for any} \quad x = G^{-1}_{i}v
				\end{align}
			hence,
				\begin{align}
					v = \bm{G} \bm{G}^{-1} v	
				\end{align}

			If we limit ourselves to allow each generative focus matrix $G_i$ to only contain $0$ or $1$, 
			and to only be formed by a combinatorial basis shuffling of the identity such that $\bm{G} \bm{G}^{-1} = I$, 
			it can be implied that each individual $G_i$ must not intersect one another.  
			From here, we can see that there are at least as many settings as the factorials of the dimension of $v$.
				 		
			\label{example_01}
		\end{example}		

		\begin{example}
			As an alternative of Example \ref{example_01}, if we fix $G$ to $I$, 
			the number of valid settings depends on the choice of individual $W_i$ that satisfies $\bm{W}' \bm{W}v = v$. 
			The task is left for us to choose the bases of each individual $W_i$ such that they are not correlated with those of the other components.
		\end{example}

		\begin{example}
			\label{pooling_example}

			In a convolutional neural network with a pooling layer \cite{lecun1998gradient},
			the generative focus is always the residue of the pooling layer that satisfies variance preservation.
			\begin{align*}
				W \Conv v \xrightarrow{pool} ( h, g )
			\end{align*}
			The generative focus here indicates from where the pooling result has taken the input.
			If $W$ has the inverse $W'$ such that $W' \Conv W = \delta $, then we can have
			\begin{align*}
				v = W' \Conv (\bm{G} \bm{G}^{-1} ( W \Conv v ) )
			\end{align*}
			where for each component $i$
			\begin{align*}
				h_i = G_i ( W \Conv v )
			\end{align*}
			The matrix $G_i$ contains the spatial information of $g$.

		\end{example}

		For a multilayer implementation of the model, 
		the mechanic of components and focuses should stay at the top of the stack as the executive function that controls thought. 
		We let the lower layers to act as the non-linear transfer function between the bottommost visible states and the topmost hidden states in the components.
	
	\subsection{Learning while executing}
			
		Thoughts are creative, and yet no one but ourselves can teach us to think with nothing but executing input sequences as the examples.
		A good thinking model should allow learning while executing.	 
		

		Like other machine learning paradigms, the model works in two phases: executing and training. 
		The difference is in our model these phases both use the same execution path with sample sequences of visible states as the only inputs.
		The system that implements the model should learn to generate the sequences and also generalize them.
		We are allowed, however, to devise specials of such sequences especially for the sake of training.

		We impose that any machine can learn while executing if 
		\begin{inparaenum}[a)]
			\item the learning happens along the path of executing, and
			\item the learning happens in the direction of executing.		
		\end{inparaenum}			
		These are the conditions for learning while executing.
		This type of training prohibits more than one step of the back propagation through time algorithm \cite{robinson1987utility} 
		that is used to train recurrent neural networks. 
		Though our model could definitely receive benefit from having some kind of a long-term guide especially for training the selective focus.	


		During the training phase of our model, we advice to train first the content parameters of the components. 		
						
		\subsubsection{Content training}	

		The content mapping parameters of each component can be trained unsupervisedly 
		with data supplied by an initial selective focus and a generative focus
		and specially designed visible state sequences to leverage them. 
		

		Let us consider a linear executing step of the model.
		\begin{align}
				v &= \bm{G} \bm{W}' \bm{W} \bm{S} \pmb{v}	\quad \forall v  \label{linear_flow}
		\end{align}
		In the manner similar to Equation \ref{linear_combine},
		$\bm{S}$ is a collection of the selective focuses of all components.
		$\pmb{v}$ represents a collection of past visible states.
		\begin{align}
			\intertext{Also for each component according to Equation \ref{detail},}
				S_i \pmb{v} &=  W'_i W_i S_i \pmb{v} \quad \forall i, \pmb{v} \label{linear_detail}
			\intertext{Merging executing equations yields}
				v &= \bm{G} \bm{S} \pmb{v} \label{equation_focus}
		\end{align}	
		Equation \ref{equation_focus} suggests how to design the training sequences, i.e.,
		the next step visible state $v$ in the sample sequence
		should somehow represent the aggregation of all selected information from the past $S \pmb{v}$ according to $G$.
		
		Given a focus $f = (s,g)$, each component will passively generate a hidden state with the current parameter values 
		and then produce a visible state. The process of learning can utilize this path:
		\begin{align*}
			S\pmb{v} \xrightarrow{tie} WS\pmb{v} = h \xrightarrow{tie} W'h = G^{-1}v
		\end{align*}
		$a \xrightarrow{tie} b$ denotes a supervised learning that learns mapping from $a$ to $b$, following its direction.
		We can see that learning of the content parameters do not break the learning while executing conditions.
						
		\subsubsection{Focus training}		
		
		This section shows that there is a possibility to train our model's focuses while satisfying the learning-while-executing conditions.	

		The generative focus is straightforward to train since it can be obtained straightforwardly by content optimization. 
		Equation \ref{equation_focus} implies that we have to choose the generative focus $G$ of all components
		such that their combined content best matches $v$, fulfilling the equation.
		Or in a convolutional network, Example \ref{pooling_example} shows that the
		generative focus $g$ of a component can be extracted along with the content while training the content parameters. 
		Then we can immediately tie the recent thoughts with it.
		\begin{align*}
			v_{t}, v_{t-1}, f_{t-1}, \ldots, v_{t-\tau}, f_{t-\tau} \xrightarrow{tie} g
		\end{align*}

		Once the content parameters and the generative focus have been trained,
		the selective focus can be trained by activate-and-tie mechanism with optionally reinforcement learning as a guide \cite{sutton1998reinforcement,zaremba2015reinforcement}.
		If the randomly chosen selective focus can allow the model to predict the next step visible state in the given sample sequence,
		according to Equation \ref{equation_focus}, the activation of that focus is subsequently enhanced.
		We will show in the next section that with the presence of memory mechanism this can be made easier.
				
	
	\subsection{Extension}

		\subsubsection{Memory component}
		\label{section_extension_memory}

		Memory is the actual implementation of a component to allow its selective focus to fetch a visible state from the past.
		The selective focus can either be in the form of 
		\begin{inparaenum}[a)]
			\item a temporal cue that contains the location in time relative to the present, or
			\item a part of a previous hidden state that allows us to fetch the residue information of that state.  
		\end{inparaenum}	
		In the latter case, the memory acts as a most-recent-hash that only allows the latest content associated with a cue the be retrieved.
		Although, a variance filling system such as the belief network, mentioned in Section \ref{section_memory_implementation}, would serve this purpose,
		this nature of the model grants us the option to use other types of memory implementation such as the exact hash table, which is free from the iterative-based training paradigm.
		The memory does not have to generalize to unseen hidden states; it only has to maintain the association between the given hidden states and their residue information,
		and we let the lower distributive representation system to handle the generalization of the focus mechanism and thought generation.

		When the selective focus is a part of some hidden state, we can use this fact to help training it.
		\begin{example}
			During a focus training phase, we can design a sample sequence so that 
			the focus of the immediate previous step can be trained one-by-one.
			Given an initial selective focus $s'$ that always points to the current visible state, 
			we can derive a hidden state that contains the information of the supposed selective focus for the previous step together with its residue:
			\begin{align*}
				v_t \xrightarrow{}_{s'_t} s_{t-1}|r
			\end{align*}
			We can then tie past states to the produced focus:
			\begin{align*}
				v_{t-1}, v_{t-2}, f_{t-2}, \ldots, v_{t-\tau}, f_{t-\tau} \xrightarrow{tie} s_{t-1}
			\end{align*}
		\end{example}


		
		\subsubsection{Context component}
		\label{section_context}

		Section \ref{temporal_section} discusses the possibility of having an alternative representation 
		model to pass on temporal conditions between steps while satisfying internal consistency. 
		This suggests that sometimes the model must be allowed to carry extra degrees of freedom 
		between steps of a thought sequence in the form of \emph{contexts}. 
		This is in order to represent the dynamic of some applications,	
		to call a few, generating a sequence of music with a fixed theme,
		modeling a world object that changes its appearance while moving, or
		addressing the transformation part of the environment according to Newton's postulation.

		Instead of allowing direct recursive links between steps, we can use components to pass on contexts.
		We can consider a \emph{context component} as a special component whose selective focus directly transforms a visible state to the component's content.
		For example, the model can cache the visible state of the immediate previous step in a context component for current use.

		Despite the flexibility provided by contexts, there can be a model without context which is, 
		in terms of complexity, equivalent to the context counterpart.
		When a context and its origin visible state are the alternative representation of each other,
		we can always replace a context component's direct transformation with a normal selective focus mechanism 
		that fetches context-equivalent data from predefined locations in a thought sequence that hold them.
		We shall use this fact to facilitate the elaboration of the proof of our model's complexity. 

						
		\subsubsection{External inputs}
	 	
		In some applications, inputs of the model are not only those generated in the sequence but also those received externally during the execution.
		Cabessa \emph{et al.} presents a theoretical framework of Super-Turing machines \cite{cabessa2012computational}.
		They are interactive Turing machines capable to handle external inputs during program execution. 
		We, on the other hand, do not treat external inputs as separated entities but rather 
		a part of visible states that are generated by external mechanisms.
		The augmented visible state $(v, u)$ therefore comprises of the part $v$ that is generated by the model and the other part $u$ that is written onto the state by the environment. The augmented part should seamlessly work with the focuses the way the normal visible state does.


\section{The model as a universal simulator}
\label{section_turing}

	We use computers to achieve much, ranging from calculating the total of a shopping cart to putting men on the moon.
	They are also potent to be used as simulators, 
	simulating the trajectories of robots or computing the motion of planets, for example,
	with limited versatility but the accuracy not less than that of our brains. 
	Because in theory we can regard computers as universal Turing machines \cite{hennie1966two}, 
	perhaps to prove that our model can sustain the thought process 
	it is to show that the model can simulate any Turing machine as such. 
	

	There have been many efforts to relate neural networks to Turing machines.
	Take Siegelmann and Sontag's work  \cite{siegelmann1991turing}, for example,
	as one among the originally cited. And during the time we compose this work, 
	two of such stand prominent among the hype in deep learning.  
	Graves \emph{et al.} presented neural Turing machines, 
	which were carefully designed using Long short-term memory (LSTM) \cite{hochreiter1997long, gers2000learning}, 
	and were tested to complete a few algorithmic tasks.
	Zaremba and Sutskever further extended the work with various of techniques, 
	notably reinforcement learning, to address the mechanism of Turing machine heads \cite{zaremba2015reinforcement}.


	Turing machine heads are comparable to our selective focuses in a sense
	when considered the ability to fetch and utilize information from the past.
	Our model also bears a resemblance to LSTM in this very aspect. 
	While Graves \emph{et al.} used LSTM to build Turing machines from neural networks, 
	the work is not a proof that LSTM by itself can simulate any Turing machine, 
	but rather the use of LSTM and other neural networks to implement Turing machine components that could. 
	This paper, starting from the concept of alternative representation, 
	develops the underlying theory that guarantees the generalization to unseen inputs,
	and integrates the concept of focus to allow the model to manipulate information in a way that resembles a Turing machine's behavior. 
	If we show that our model can simulate any Turing machine, 
	we could say that this work bridges the gap between LSTM and Turing machines, 	
	providing another evidence that perhaps a neural network can also be viewed as a universal Turing machine.

	Before we go to our proof, we consider another simple lemma that relates Turing machine's symbols with our state representations.
	\begin{lemma}
		There always exists a set of alternative symbols for a Turing machine's transition function, 
		\begin{align*}
			(s,t_h) \to (s',t'_h,h')
		\end{align*}
		that satisfies the non-sharing property from any current state $s$ and the tape content $t_h$ associated with any current head $h$ to
		a new state $s'$, a new head location $h'$, or a new content $t'_h$ to be written over the current head.
		\label{lemma_alphabet}
	\end{lemma}
	Since the transition function of a Turing machine is one-directional and deterministic, we can readily implement it in a system with the non-sharing property.
	Proposition \ref{lemma_combination} suggests that we can potentially use a multi-layer implementation to generatively map any current state and the tape content associated with any current head to a new state, a new head location, and a new content to be written over the current head.
	Here the lemma certifies that we can always find a set of alternative symbols that satisfies Equation \ref{combine}, 
	and now we are ready to show example algorithms for simulating a Turing machine on our model with, of course, their complexity analysis.

	\subsection{An algorithm}
	
	When running a program on a Turing machine, the dynamic of the program during the execution time 
	prevents us from diverting the focus to directly point to the Turing machine's head location. 
	We can resort to search for the current head content, which is generatively produced somewhere in the visible state sequence.
	At the start of each Turing machine step, using only a fixed number of recent thoughts
	the algorithm makes the model looks back into the sequence to resolve the current head content.

	Consider a Turing machine's transition function, $(s,t_h) \to (s',t'_h,h')$.
	Since our original model does not have memory, it is required to write down all the symbols onto the visible state sequences.
	Let each square bracket represents a visible state of our model's sequence.
	Here is a portion of the sequence involved in one Turing machine step:
	\begin{align*}
		\left[ \ldots, m_{-1} \right], 
		\left[ \beta, s, h, (t_{h_{-1}}, h_{-1}), t_{h_{-1}}, \ast, \ast \right], 
		\left[ \alpha, s, h, \ast, h_{f_{-1}}, t_{h_{f_{-1}}}, f_{-1}, 1 \right], 
		\ldots, 
		\left[ \alpha, s, h, \ast, h, t_{h}, f_{-m}, m \right],
	\end{align*}
	The sequence comprises of a pivot step, tagged with $\beta$,
	and several search steps, tagged with $\alpha$.
	At the pivot step, $s$ is the current state, and $h$ is the current head.
	The parenthesis in the pivot groups the new content of the previous head location with the head itself. 
	Next in the step is the yet-to-be-resolved $t_{h_{-1}}$ content for the current head.
	To resolve this value, the search steps will jump from one pivot step to another looking for the head's content in the parentheses.
	Each search step, starting with $\alpha$, carries $s, h$ to be used to compute the symbols of the next Turing machine step at the end of the search.
	The search progresses by updating these four parameters: the found head, the content of the found head, 
	the focus location of the found head, and the step counter of the search.
	The model keeps track of the found head to conditionally decide when to end the search.
	The focus location allows the model to track the current search location.
	The step counter allows the model to compute how many steps to jump to find the next pivot step.
	And it is already obvious why we keep the found head content.
	These four parameters allow the model to evaluate the selective focus using only the two most recent visible states.
	$\ast$ represents a wild card symbol which we do not interest at the time it depicts.
	We encourage the reader to become familiar with this sequence before moving forward.

	To generate a new visible state, the model utilizes components.  
	Each individual component copies information from the sequence following the focus,  
	satisfying internal consistency, and combines with that of the others to generate symbols (Lemma \ref{lemma_alphabet}).
	For the sake of simplicity, we also use context components on this algorithm's illustration.
	It is not hard to see that the information keeps in each context component satisfies the non-sharing property. 
	Therefore, we can entirely replace each context with some initial symbols at the very beginning of the sequence 
	and a focus mechanism to fetch and combine them, without affecting the algorithm's performance (Section \ref{section_context}).
	At each search step (with $\alpha$), the model decides whether to stop or continue the search and puts symbols on the components.
	The following describes the templates of what information will be carry in the components:
	\begin{align*}
		\left[ \alpha, s, h, t_{h_{-1}}, h_{f_{-m}}, t_{h_{f_{-m}}}, f_{-m}, m \right] \to
		\begin{array}{c}
			\text{step tag (context)}		\\
			\text{Turing machine' states (copy)}			\\
			\text{current search head (copy)}  			\\
			\text{current search content (copy)} 			\\
			\text{current search focus (context)} 			\\
			\text{search iteration (context)} 							
		\end{array} \to \text{new visible state}
	\end{align*}
	i.e., when $h_{f_{-m}} \in \{h, \phi \}$ the templates collect these symbols:
	\begin{align*}
		\begin{array}{c}
			\beta		\\
			s, h		\\
			h_{f_{-m}}  \\
			t_{h_{f_{-m}}} \\
			\ast \\
			\ast 							
		\end{array} = 
		\begin{array}{c}
			\beta		\\
			s, h		\\
			h  \\
			t_{h} \\
			\ast \\
			\ast 							
		\end{array} &\to
		\left[ \beta, s', h', (t'_{h}, h), t_{h}, \ast, \ast \right] \to \ldots
		\intertext{ otherwise,}
		\begin{array}{c}
			\alpha										\\
			s, h										\\
			h_{f_{-m-1}}								\\
			t_{h_{f_{-m-1}}}								\\
			f_{-m-1} = f_{-m} - m_{-m} - 1				\\ 
			m + 1							
		\end{array} &\to
		\left[ \alpha, s, h, \ast, h_{f_{-m-1}}, t_{h_{f_{-m-1}}}, f_{-m-1}, m+1 \right] \to \ldots
	\end{align*}	
	$\phi$ represents an empty or any unrecognized symbol at the start of the sequence.
	For the first search step after a pivot step,
	\begin{align*}
		\left[ \ldots, m_{-1} \right], \left[ \beta, s, h, (t_{h_{-1}}, h_{-1}), t_{h_{-1}}, \ast, \ast \right] \to
		\begin{array}{c}
			\alpha						\\
			s, h						\\
			h_{f_{-1}}					\\
			t_{h_{f_{-1}}} 				\\
			f_{-1} = - m_{-1} - 1		\\
			1							
		\end{array} \to
		\left[\alpha, s, h ,\ast, h_{f_{-1}}, t_{h_{f_{-1}}}, f_{-1}, 1 \right] \to \ldots
	\end{align*}		
	$f_{-1}$ is fixed to the previous pivot step.
	The components we show here are merely the templates of the real components. 
	They do not correspond injectively to the symbols on the visible state, 
	rather each symbol on the visible state derives from a mapping from some of the symbols presenting in them.
	For example, the new state $s'$ would require the old state and the found head content to be generated.
	More importantly, the reader can verify that the identities of the symbols in each component template 
	can be determined within at most two most recent visible states.
	Since the first visible state in the sequence has to be given as the input,
	this completes the induction for each transformation step of a Turing machine.

		\subsubsection{Complexity analysis}	
			
			Let $N$ be the current number of Turing machine steps counting from the beginning of program execution.
			The worst case time complexity of the algorithm to execute the step, 
			contributed by the search procedure, are bounded by $O(N)$,
			assuming that each step is required to search to the begin of the sequence.
			
			What is the average time complexity to execute a step?
			
\begin{figure}[h!]
		\centering
		\def\svgwidth{0.5\textwidth}
		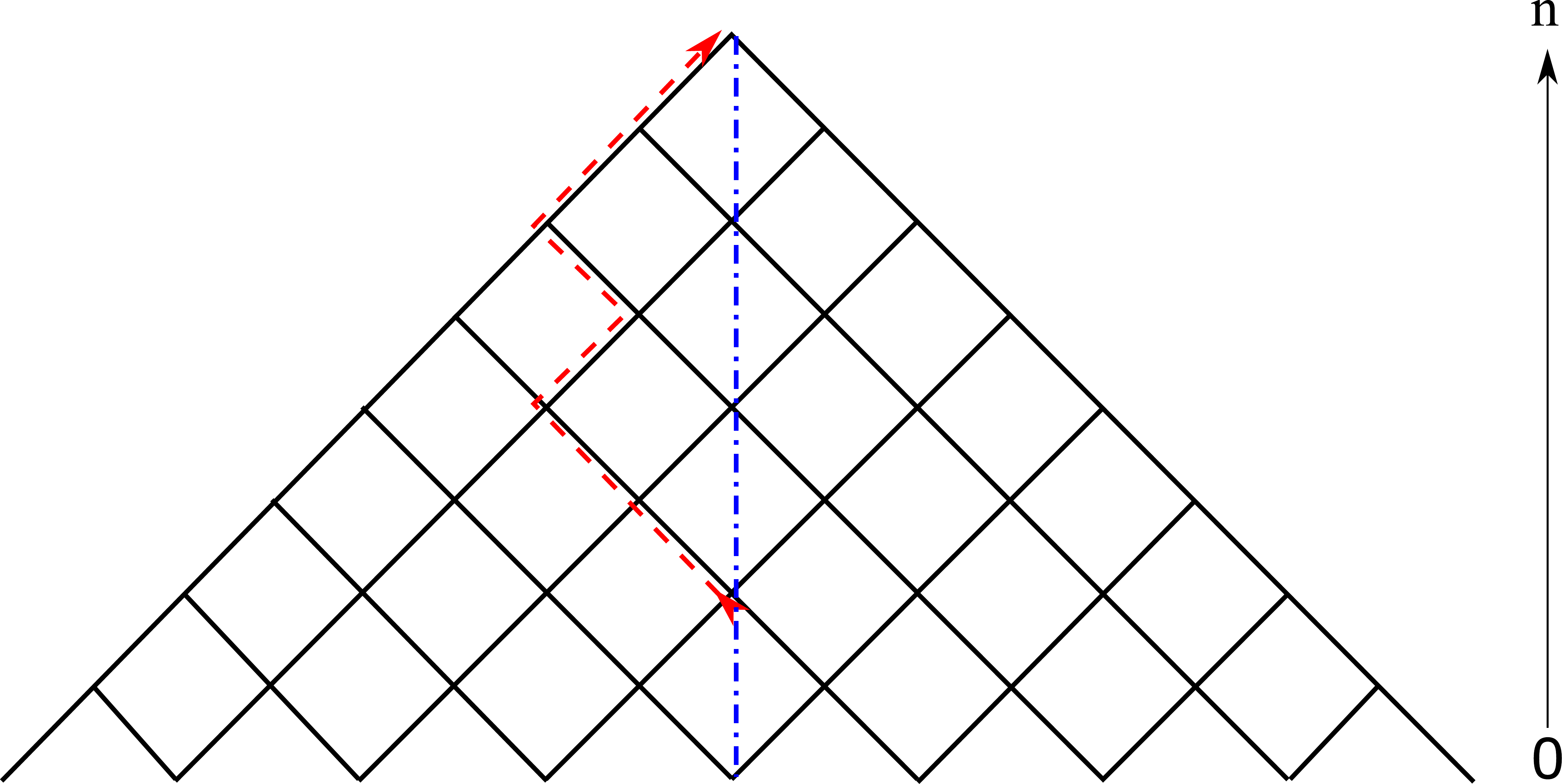
		\caption{A graph of all possible paths a Turing machine's head can take to a location.
		The red dash-dash line illustrates a path from the step $n = 2$ and never again reaches the location until the current step. }
		\label{random_walk}
	\end{figure}

			For a vanilla Turing machine, the head only moves left or right. 
			Under the assumption that there is an equal probability for the head to move left or right 
			considering all possible programs, we can plot a random walk graph of the head's locations
			up until the head reaches the current location (see Figure \ref{random_walk}). 

			Consider the head's location of a Turing machine after $N$ steps of execution, 
			the total number of paths to that point is equal to
			$2^N = |P_{\infty}| + |P_{0}| + |P_{1}| + \ldots + |P_{N-2}|$ 
			where $|P_n|$ is the number of paths where the previous visits of the current location were at the $n$-th step, 
			$|P_\infty|$ is the number of paths that never visit the location.
						
			The expected total of the search steps is given by
			\begin{align}
				E\left[\text{search steps at }N\right] &= \left[ \frac{1}{2^N}\sum^{N-1}_{n = 0} (N - n) |P_n| \right]  + \left[ \frac{N}{2^N}\left(2^N - \sum^{N-1}_{n = 0} |P_n| \right) \right] \label{bound} 
			\end{align}		
			The first term in Equation \ref{bound} is the expected number of steps from the last visit of the current location. 
			$|P_n|$ can be computed by counting the number of paths, on the left and the right of the Pascal's semi-triangle, 
			avoiding the current location (see Figure \ref{random_walk}). It is given by the Catalan number:
			\begin{align*}
				|P_n| &= 2 C(N - n-2, 0) 2^n \\
				C(2m, 0) &= \frac{1}{m+1} \binom{ 2m }{ m }\\
				C(2m+1,0) &= 0 \\
				\intertext{using Stirling approximation on the Catalan number, we can compute the asymptote:}
							|P_n| &= O\left(\frac{2^N}{(N-n)^{3/2}}\right)						
			\end{align*}
			
			The asymptote of the first term is given by $O( \sqrt{N} )$:
			\begin{align*}			
				\frac{1}{2^N} \sum^{N-1}_{n = 0} (N-n) |P_n| &= \sum^{N-2}_{n = 0} 2 (N-n) C(N-n-2, 0) \frac{2^n}{2^N}	 \\
				&= \sum^{N-2}_{n = 0} O\left(\frac{1}{(N-n)^{1/2}}\right)
			\end{align*}	
			Because $\frac{1}{(N-n)^{1/2}}$ is monotonically increasing, we can find its bounding via the integral test:
			\begin{align*}	
				\int^{N-1}_0 \! \frac{j}{(N-n + 1)^{1/2}} \, \mathrm{d}n &< \sum^{N-2}_{n = 0} O\left(\frac{1}{(N-n)^{1/2}}\right) < \int^{N-1}_0 \! \frac{k}{(N-n)^{1/2}} \, \mathrm{d}n \quad \exists j, k \in \mathbb{R}^+\\
				-2j(N-n+1)^{1/2} \bigg\vert_{0}^{N-1} &< \sum^{N-2}_{n = 0} O\left(\frac{1}{(N-n)^{1/2}}\right) < -2k(N-n)^{1/2} \bigg\vert_{0}^{N-1} \\
				O(\sqrt{N}) &< \sum^{N-2}_{n = 0} O\left(\frac{1}{(N-n)^{1/2}}\right) < O(\sqrt{N}) 
			\end{align*}		
			
			The second term in Equation \ref{bound} accounts for the paths where the current head location has never been visited before. In this case, the model has to search the entire sequence yielding the asymptote of $O(N)$ for the term. Fortunately, we can make a modification to the algorithm to get rid of this term by adding two extra copy components that keeps track of the leftmost and rightmost bounds of all the visited head locations. When the head location exceeds one of the bounds, the model updates the bound, skips the search and immediately writes the empty symbol as the head's content. This augmentation improves the average time complexity of the algorithm to $O(\sqrt{N})$.

			Can we do better?
						
	\subsection{A constant time algorithm}

	\begin{figure}[ht!]
		\centering
		\def\svgwidth{0.4\textwidth}
		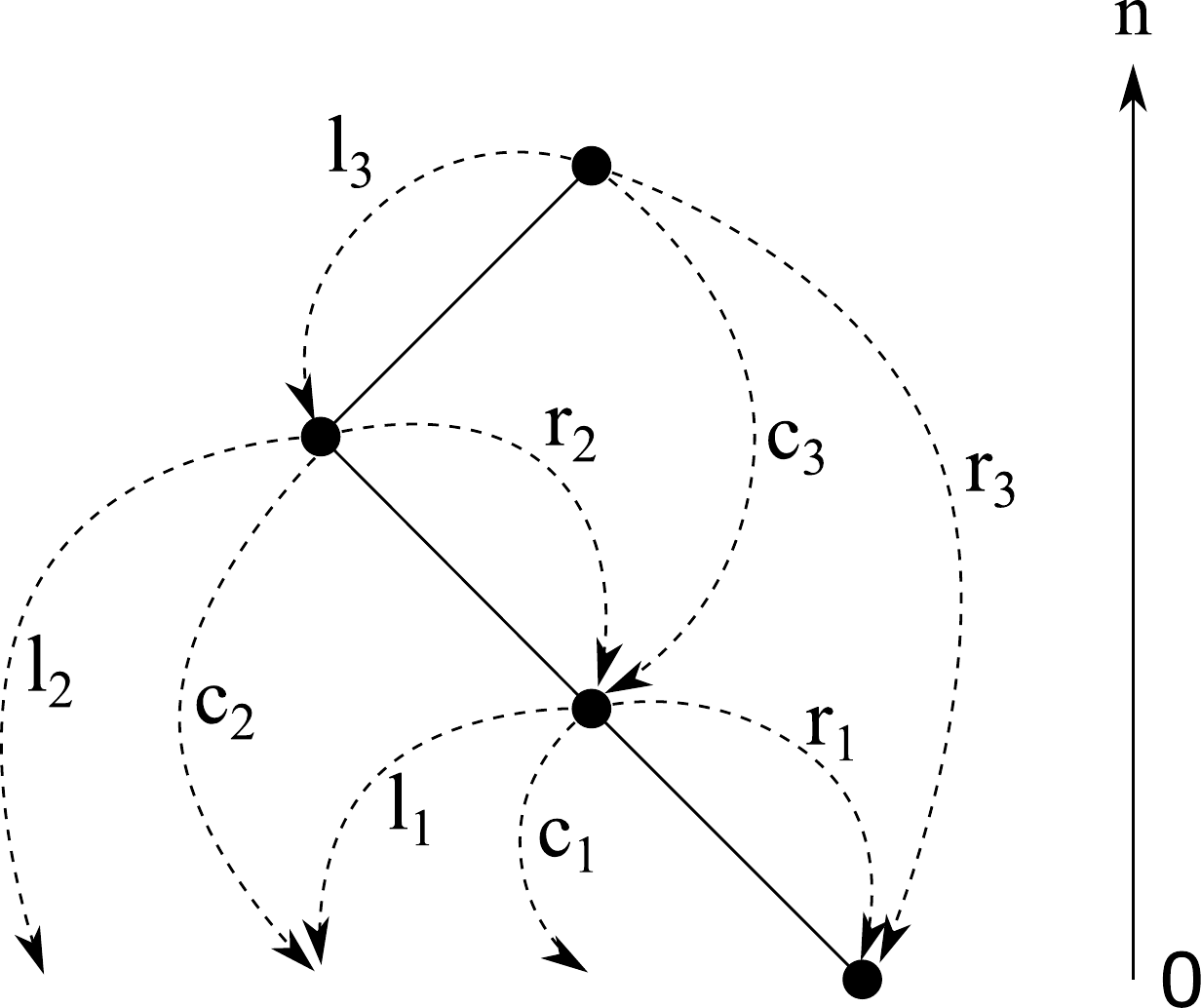
		\caption{An illustration of a Turing machine's head location from start after moving left, right, right. 
		The dashed arrows indicate the number of steps from the current to the previous left of the current head location, to the previous right of it, or to itself.}
		\label{best_algor}
	\end{figure}

		The idea of this algorithm is to let every step maintains the step counts from the current
		to the previous left of the current head location, to the previous right of it, and to itself.
		Let us consider now three steps of a visible state sequence from the start of a program.
		From step $n = 0$, the head moves left, left, then right, producing the following sequence:
		
		\begin{align*}
			\begin{array}{cccccccccc}
				\left[\right. \text{Tag}, & s, & h, & t_h, & t'_h, & d_{n-1 \to n}, & l_n, & c_n, & r_n &\left.\right] \\
				& & & & & & & &	& \\
				\left[\right. \beta 	,& s_0	,& h_0	,& t_{h_0} 	,& \ast 	,& \ast 	,& 0 	,& 0 	,& 0 	&\left.\right], \\
				\left[\right. \gamma  	,& s_1	,& h_1	,& \ast		,& t'_{h_0} ,& L 		,& \ast	,& \ast	,& \ast &\left.\right], \\
				\left[\right. \alpha  	,& \ast ,& \ast	,& \ast 	,& \ast 	,& \ast		,& \ast	,& -1 	,& -1 	&\left.\right], \\
				\left[\right. \beta 	,& s_1	,& h_1	,& t_{h_1} 	,& \ast 	,& \ast 	,& -1 	,& -1 	,& -1 	&\left.\right], \\
				\left[\right. \gamma  	,& s_2	,& h_2	,& \ast		,& t'_{h_1} ,& L 		,& \ast	,& \ast	,& \ast &\left.\right], \\
				\left[\right. \alpha  	,& \ast ,& \ast	,& \ast 	,& \ast 	,& \ast		,& \ast	,& -2 	,& -1 	&\left.\right], \\
				\left[\right. \beta 	,& s_2	,& h_2	,& t_{h_2} 	,& \ast 	,& \ast 	,& -2 	,& -2 	,& -1 	&\left.\right], \\
				\left[\right. \gamma  	,& s_3	,& h_3	,& \ast		,& t'_{h_2} ,& R 		,& \ast	,& \ast	,& \ast &\left.\right], \\
				\left[\right. \alpha  	,& \ast ,& \ast	,& \ast 	,& \ast 	,& \ast		,& -1	,& -2 	,& \ast &\left.\right], \\
				\left[\right. \beta 	,& s_3	,& h_3	,& t_{h_3} 	,& \ast 	,& \ast 	,& -1 	,& -2 	,& -3 	&\left.\right], \\
				\left[\right. \ldots 	&\ldots  &\ldots &\ldots 	 &\ldots 	 &\ldots 	 &\ldots &\ldots &\ldots &\left.\right],			
			\end{array}
		\end{align*}
		$d_{n-1 \to n}$ represents a head direction from step $n-1$ to $n$.
		The previous step counts for the previous left, right, and the current position are $l_n$ $r_n$ $c_n$ respectively.
		A Turing machine's step starts at $\beta$ with its states and the previous step counts.
		At $\gamma$ step, the next Turing machine's states are produced.
		At $\alpha$ step, the previous step counts to the current head location and one of left or right are produced depending on the head direction.
		At the next $\beta$, our model finishes the Turing's step with the fetched head content and the last step count. 
		The generation of the step counts follows these rules,
		\begin{align*}
			c_n &= 
			\begin{cases}
				l_n - 1 & \mbox{if } d_{n-1 \to n} = R \\
				r_n - 1 & \mbox{otherwise}
			\end{cases} \\
			l_n &= 
			\begin{cases}
				c_n + l_{(n + c_n)} & \mbox{if } d_{n-1 \to n} = L \\
				- 1 & \mbox{otherwise}
			\end{cases} \\
			r_n &= 
			\begin{cases}
				c_n + r_{(n + c_n)} & \mbox{if } d_{n-1 \to n} = R \\
				- 1 & \mbox{otherwise}
			\end{cases}
		\end{align*}
		The component templates are as follow: 
		\begin{align*}
			\left[ \text{Tag}, s, h, t_{h}, t'_h, d_{n-1 \to n}, l_n, c_n, r_n \right] \to
			\begin{array}{c}
				\text{step tag (context)}		\\
				\text{Turing machine' states (copy)}			\\
				\text{previous head content (copy)} 			\\
				\text{head direction (context)} 			\\
				\text{step counts (context)}				
			\end{array} \to \text{new visible state}
		\end{align*}
		Again, the reader can verify that the identities of the symbols in each component template
		can be determined within this time the most recent visible state. This completes the algorithm. 
		And because each Turing machine's step only requires two extra steps to be simulated in our model. 
		We can conclude that the time complexity of this algorithm is $O(1)$.

	The existence of this algorithm by itself allows us to state the main theorem of this paper.
	\begin{theorem}
		The model with an arbitrary depth and a finite number of components can simulate any vanilla Turing machine 
		and only be slowed by within a constant factor compared to the machine it simulates.
	\end{theorem}

	\subsection{With memory components}
	
		We can augment the components with the most-recent-hash memory mechanism like one we introduced in Section \ref{section_extension_memory}.
		This enhancement allows the tape content associated with any head location,  
		not just ones limited by the left-right-only movement of the vanilla Turing machines, to be read or written in a single step.
		An execution step of a Turing machine simulated on our model with the most-recent-hash memory is manifested by these expressions:
		\begin{align*}
			\left[ \beta, h, \ast, s \right] \to
			\begin{array}{c}
				\alpha \\
				h \xrightarrow{read} t_h \\
				s
			\end{array} \to 
			\left[ \alpha, h', t'_h, s' \right] \to
			\begin{array}{c}
				\beta \\
				t'_h \xrightarrow{write} h \\
				s'
			\end{array} \to 
			\left[ \beta, h', t'_h, s' \right] \to \ldots
		\end{align*}
		For each Turing machine's step, our model only needs one extra step to complete the read-write cycle
		with one memory component to read and write the tape contents.
		It infers the head content $t_h$ at the $\beta \to \alpha$ step
		and memorizes the new head content $t'_h$ at the $\beta \to \alpha$ step, 
		both while focusing on the head symbol at the $\beta$ step.
		The efficiency is readily apparent.


\section{Discussion}
	
	To state it one last time, this paper presents a theoretical thinking model.
	The model consists of components where all combine the information that each fetches 
	from some part of the former thoughts in order to creatively compose a new one.
	Each component has the ability to invariantly extract information from any when and where with its selective focus,
	which in turn depends on and is driven only by the most recent thoughts.
	The combination mechanism is governed by the concept of internal consistency.
	Internal consistency is especially useful for a system that operates through time,
	has the capacity to alternatingly and repeatedly recognize and generate data, 
	and requires that the newly generated data are relevant to the cause from which they are originated, such as the thinking process. 

	The explicit use of the focuses brings an advantage to our model.
	It happens that, in our world, physical interaction of between objects tend to correlate their spatio-temporal locations. 
	Because of this, physical mechanic can be hypothetically simulated using a hierarchy of execution within a system.
	Our model allows the focus mechanism to handle a higher order of execution,
	while having each component handles the transformation within its responsible detail.
	This way we might be able to achieved generalization with limited training data.
	It is also a means to reduce the hypothesis variance of the model's implementation 
	while, as much as possible, preserving the degrees of freedom of thinking. 

	The problem this model tries to solve is in fact the inverse of filtering in signal processing.
	In filtering we try to discover a recognition function, or simply an estimate in the context of signal process,
	while taking the generative function, or the input control, and process noise into account. 
	In this work however, we attempt to find the generative function that is simply the perfect inverse of the recognition function.
	Furthermore the approach of filtering has also been widely applied to find distributed weights 
	for combining series of measurements that signify the same information
	proportionally to each measurement's confidential quantity.
	While in our model however, the distributed weights are the mapping from hidden states to visible states,
	and are formed by training with the only limitation to preservation of variance.

	Throughout the length of this work, we have discussed several models.	
	The first model that we introduce in Section \ref{section_internal_consistency} is the hidden-visible bipartite model
	with the hidden and visible state representations that are always the alternative representations of each other.
	Applications that involve this basic model include factor analysis and some knowledge representation where preservation of variance is required.
	Adding the temporal support to the first model gives rise to the temporal version of it (Section \ref{temporal_section}).
	The possibility to further enhance the temporal model with indefinitely long memory retention
	has eventually led the discussion to our thinking model in Section \ref{section_model}. 
	We dedicate Section \ref{section_turing} to show that the model can simulate any Turing machine 
	with the computational complexity rivaling that of a universal Turing machine.
	Table \ref{table_models} summarizes all the models we have discussed in this work.
	\begin{table}[h]
		\center
		\begin{tabular} {|l|l|l|l|}
			\hline
			Model & Hidden state & Visible state & complexity \\
			\hline
			Hidden-visible bipartite 	&	fixed					& fixed								&	bidirectional mapping \\
			Temporal bipartite			&	changed by contexts 	& changed by contexts				&	recurrent model \\
			Thinking model				&	changed by focuses		& contents follow focuses		&	universal Turing machine	\\
			\hline
		\end{tabular}
		\caption{Summary of the models discussed in this paper.}
		\label{table_models}
	\end{table}

	To make an intelligence machine with the generative capability, 	
	one essentially requires a decent way to internally represent the world.
	Deep learning and techniques such as sparse representation are particularly designed to address this requirement.
	In this paper, we present another important factor that allows generalization guarantee of newly generated data to be ``relevant'',
	at least to all the data accumulated during the course of the machine's execution.
	Internal consistency is not merely a hypothetical trait of intelligent machines when they manipulate data. 
	As in humans, we believe that when we factorize knowledge into parts
	to identify the similarity and distinction in the information
	and to be able to combine with other knowledge to form a new idea is when we truly understand something.
	This conviction serves as the very motive of this work.

\bibliographystyle{unsrt}
\bibliography{references}

\begin{thebibliography}{10}

\bibitem{boole1854investigation}
George Boole.
\newblock {\em An investigation of the laws of thought: on which are founded
  the mathematical theories of logic and probabilities}.
\newblock Dover Publications, 1854.

\bibitem{turing1950computing}
Alan~M Turing.
\newblock Computing machinery and intelligence.
\newblock {\em Mind}, pages 433--460, 1950.

\bibitem{hinton2006fast}
Geoffrey~E Hinton, Simon Osindero, and Yee-Whye Teh.
\newblock A fast learning algorithm for deep belief nets.
\newblock {\em Neural computation}, 18(7):1527--1554, 2006.

\bibitem{bengio2009learning}
Yoshua Bengio.
\newblock Learning deep architectures for ai.
\newblock {\em Foundations and trends{\textregistered} in Machine Learning},
  2(1):1--127, 2009.

\bibitem{schmidhuber2015deep}
J{\"u}rgen Schmidhuber.
\newblock Deep learning in neural networks: An overview.
\newblock {\em Neural Networks}, 61:85--117, 2015.

\bibitem{hochreiter1997long}
Sepp Hochreiter and J{\"u}rgen Schmidhuber.
\newblock Long short-term memory.
\newblock {\em Neural computation}, 9(8):1735--1780, 1997.

\bibitem{graves2009novel}
Alex Graves, Marcus Liwicki, Santiago Fern{\'a}ndez, Roman Bertolami, Horst
  Bunke, and J{\"u}rgen Schmidhuber.
\newblock A novel connectionist system for unconstrained handwriting
  recognition.
\newblock {\em Pattern Analysis and Machine Intelligence, IEEE Transactions
  on}, 31(5):855--868, 2009.

\bibitem{hennie1966two}
Fred~C Hennie and Richard~Edwin Stearns.
\newblock Two-tape simulation of multitape turing machines.
\newblock {\em Journal of the ACM (JACM)}, 13(4):533--546, 1966.

\bibitem{koller2009probabilistic}
Daphne Koller and Nir Friedman.
\newblock {\em Probabilistic graphical models: principles and techniques}.
\newblock MIT press, 2009.

\bibitem{hyvarinen2000independent}
Aapo Hyv{\"a}rinen and Erkki Oja.
\newblock Independent component analysis: algorithms and applications.
\newblock {\em Neural networks}, 13(4):411--430, 2000.

\bibitem{goodfellow2014generative}
Ian Goodfellow, Jean Pouget-Abadie, Mehdi Mirza, Bing Xu, David Warde-Farley,
  Sherjil Ozair, Aaron Courville, and Yoshua Bengio.
\newblock Generative adversarial nets.
\newblock In {\em Advances in Neural Information Processing Systems}, pages
  2672--2680, 2014.

\bibitem{vincent2010stacked}
Pascal Vincent, Hugo Larochelle, Isabelle Lajoie, Yoshua Bengio, and
  Pierre-Antoine Manzagol.
\newblock Stacked denoising autoencoders: Learning useful representations in a
  deep network with a local denoising criterion.
\newblock {\em The Journal of Machine Learning Research}, 11:3371--3408, 2010.

\bibitem{szegedy2014going}
Christian Szegedy, Wei Liu, Yangqing Jia, Pierre Sermanet, Scott Reed, Dragomir
  Anguelov, Dumitru Erhan, Vincent Vanhoucke, and Andrew Rabinovich.
\newblock Going deeper with convolutions.
\newblock {\em arXiv preprint arXiv:1409.4842}, 2014.

\bibitem{gatys2015neural}
Leon~A Gatys, Alexander~S Ecker, and Matthias Bethge.
\newblock A neural algorithm of artistic style.
\newblock {\em arXiv preprint arXiv:1508.06576}, 2015.

\bibitem{nishimoto2011reconstructing}
Shinji Nishimoto, An~T Vu, Thomas Naselaris, Yuval Benjamini, Bin Yu, and
  Jack~L Gallant.
\newblock Reconstructing visual experiences from brain activity evoked by
  natural movies.
\newblock {\em Current Biology}, 21(19):1641--1646, 2011.

\bibitem{le2011ica}
Quoc~V Le, Alexandre Karpenko, Jiquan Ngiam, and Andrew~Y Ng.
\newblock Ica with reconstruction cost for efficient overcomplete feature
  learning.
\newblock In {\em Advances in Neural Information Processing Systems}, pages
  1017--1025, 2011.

\bibitem{oja1982simplified}
Erkki Oja.
\newblock Simplified neuron model as a principal component analyzer.
\newblock {\em Journal of mathematical biology}, 15(3):267--273, 1982.

\bibitem{nair2010rectified}
Vinod Nair and Geoffrey~E Hinton.
\newblock Rectified linear units improve restricted boltzmann machines.
\newblock In {\em Proceedings of the 27th International Conference on Machine
  Learning (ICML-10)}, pages 807--814, 2010.

\bibitem{cybenko1989approximation}
George Cybenko.
\newblock Approximation by superpositions of a sigmoidal function.
\newblock {\em Mathematics of control, signals and systems}, 2(4):303--314,
  1989.

\bibitem{atlas1988artificial}
Les~E Atlas, Toshiteru Homma, and Robert~J Marks~II.
\newblock An artificial neural network for spatio-temporal bipolar patterns:
  Application to phoneme classification.
\newblock In {\em Proc. Neural Information Processing Systems (NIPS)}, page~31,
  1988.

\bibitem{taylor2006modeling}
Graham~W Taylor, Geoffrey~E Hinton, and Sam~T Roweis.
\newblock Modeling human motion using binary latent variables.
\newblock In {\em Advances in neural information processing systems}, pages
  1345--1352, 2006.

\bibitem{newton1999principia}
Isaac Newton.
\newblock {\em The principia: mathematical principles of natural philosophy}.
\newblock Univ of California Press, 1999.

\bibitem{lecun1998gradient}
Yann LeCun, L{\'e}on Bottou, Yoshua Bengio, and Patrick Haffner.
\newblock Gradient-based learning applied to document recognition.
\newblock {\em Proceedings of the IEEE}, 86(11):2278--2324, 1998.

\bibitem{robinson1987utility}
AJ~Robinson and Frank Fallside.
\newblock {\em The utility driven dynamic error propagation network}.
\newblock University of Cambridge Department of Engineering, 1987.

\bibitem{sutton1998reinforcement}
Richard~S Sutton and Andrew~G Barto.
\newblock {\em Reinforcement learning: An introduction}, volume~1.
\newblock MIT press Cambridge, 1998.

\bibitem{zaremba2015reinforcement}
Wojciech Zaremba and Ilya Sutskever.
\newblock Reinforcement learning neural turing machines.
\newblock {\em arXiv preprint arXiv:1505.00521}, 2015.

\bibitem{cabessa2012computational}
J{\'e}r{\'e}mie Cabessa and Hava~T Siegelmann.
\newblock The computational power of interactive recurrent neural networks.
\newblock {\em Neural Computation}, 24(4):996--1019, 2012.

\bibitem{siegelmann1991turing}
Hava~T Siegelmann and Eduardo~D Sontag.
\newblock Turing computability with neural nets.
\newblock {\em Applied Mathematics Letters}, 4(6):77--80, 1991.

\bibitem{gers2000learning}
Felix~A Gers, J{\"u}rgen Schmidhuber, and Fred Cummins.
\newblock Learning to forget: Continual prediction with lstm.
\newblock {\em Neural computation}, 12(10):2451--2471, 2000.

\end{thebibliography}

\end{document}